\pdfoutput=1

\documentclass[conference]{IEEEtran}
\usepackage[bookmarks=false]{hyperref}
\usepackage{amsmath,amssymb,amsfonts}%
\usepackage{amsthm}
\usepackage{algorithm}
\usepackage{array}
\usepackage[caption=false,font=normalsize,labelfont=sf,textfont=sf]{subfig}
\usepackage{textcomp}
\usepackage{stfloats}
\usepackage{url}
\usepackage{verbatim}
\usepackage{graphicx}
\usepackage{cite}
\usepackage{tikz}
\usepackage{forest}
\usepackage{pgfplots} 

\usepackage{pgf-pie}  
\usepackage{listings}%
\usepackage{comment}
\usepackage{manyfoot}%
\usepackage{booktabs}%
\usepackage{tabularx}
\usepackage{makecell}
\usepackage{adjustbox}

\usepackage{multirow}%
\usepackage{tabularx}
\usepackage{makecell}
\usepackage{mathrsfs}%
\usepackage[figuresright]{rotating}%
\usepackage{xcolor}%
\usepackage{textcomp}%
\usepackage{algorithmicx}%
\usepackage{algpseudocode}%
\usepackage{program}%
\usepackage{longtable}
\usepackage{multirow}
\usepackage{array}
\usepackage{comment}
\usepackage{tabularx}
\usepackage{color,soul}
\def\BibTeX{{\rm B\kern-.05em{\sc i\kern-.025em b}\kern-.08em
    T\kern-.1667em\lower.7ex\hbox{E}\kern-.125emX}}
\begin{document}

\title{A Survey of Anomaly Detection in In-Vehicle Networks}

\author{\IEEEauthorblockN{Övgü Özdemir}
\IEEEauthorblockA{\textit{R\&D Department} \\
\textit{Proven Technologies Inc.}\\
Ankara, Turkey \\
oozdemir@proven.technology}
\and
\IEEEauthorblockN{M. Tuğberk İşyapar}
\IEEEauthorblockA{\textit{Department of Computer Engineering} \\
\textit{Middle East Technical University}\\
Ankara, Turkey \\
tugberk@ceng.metu.edu.tr}
\and
\IEEEauthorblockN{Pınar Karagöz}
\IEEEauthorblockA{\textit{Department of Computer Engineering} \\
\textit{Middle East Technical University}\\
Ankara, Turkey \\
karagoz@ceng.metu.edu.tr}
\and
\IEEEauthorblockN{Klaus Werner Schmidt}
\IEEEauthorblockA{\textit{Department of Electrical-Electronics Engineering} \\
\textit{Middle East Technical University}\\
Ankara, Turkey \\
schmidt@metu.edu.tr}
\and
\IEEEauthorblockN{Demet Demir}
\IEEEauthorblockA{\textit{R\&D Department} \\
\textit{Proven Technologies Inc.}\\
Ankara, Turkey \\
ddemir@proven.technology}
\and
\IEEEauthorblockN{N. Alpay Karagöz}
\IEEEauthorblockA{\textit{R\&D Department} \\
\textit{Proven Technologies Inc.}\\
Ankara, Turkey \\
akaragoz@proven.technology}
}



\maketitle

\begin{abstract}
Modern vehicles are equipped with Electronic Control Units (ECU) that are used for controlling important vehicle functions including safety-critical operations. ECUs exchange information via in-vehicle communication buses, of which the Controller Area Network (CAN bus) is by far the most widespread representative. Problems that may occur in the vehicle's physical parts or malicious attacks may cause anomalies in the CAN traffic, impairing the correct vehicle operation. Therefore, the detection of such anomalies is vital for vehicle safety. This paper reviews the research on anomaly detection for in-vehicle networks, more specifically for the CAN bus. Our main focus is the evaluation of methods used for CAN bus anomaly detection together with the datasets used in such analysis. To provide the reader with a more comprehensive understanding of the subject, we first give a brief review of related studies on time series-based anomaly detection. Then, we conduct an extensive survey of recent deep learning-based techniques as well as conventional techniques for CAN bus anomaly detection. Our comprehensive analysis delves into anomaly detection algorithms employed in in-vehicle networks, specifically focusing on their learning paradigms, inherent strengths, and weaknesses, as well as their efficacy when applied to CAN bus datasets.  Lastly, we highlight challenges and open research problems in CAN bus anomaly detection.
\end{abstract}


\section{Introduction}
\label{section:intro}
\IEEEPARstart{M}{odern} vehicles can be regarded as complex Internet of Things (IoT) devices \cite{song2016intrusion}, equipped with multiple Electronic Control Units (ECU) \cite{song2021self} that communicate through in-vehicle networks. ECUs connect sensors and actuators via constantly transmitting and receiving data containing vehicle parameters that vary from engine functions and other sensory information to entertainment and convenience features. Several standard communication buses constitute the medium of communication, among which, Controller Area Network (CAN) \cite{CANstandard} is the most extensively used message-based protocol. 

CAN bus was developed by Bosch in the mid-1980s to enable in-vehicle communication. Unlike other communication networks such as switched Ethernet, CAN broadcasts messages to the entire network instead of forwarding them from one point to another. CAN bus allows only one message to be transmitted at a time. A unique identifier (ID) in the CAN message determines the priority of the message in traffic. The transmitted message reaches all nodes and a node itself decides to either receive the message or ignore it. Each CAN message carries data from different sources such as sensors and actuators. Depending on the importance of the enclosed signals, the transmission frequency of these messages on the CAN bus can vary. These messages can be further associated with the receive time as the message timestamp.

Continuously evolving CAN bus data contain valuable information about the status of the vehicle, which can be exploited to extract patterns to predict and forecast the future of the equipment \cite{theissler2021predictive}. Information extraction at scale requires data-oriented methods to come into play since static rule-based specifications easily get over-complicated and become incompetent as data volume soars. 

Vehicles, being complex structures, are prone to anomalies, with the origin of anomalies not being trivially identified due to numerous interacting parts in rapidly changing conditions \cite{theissler2021predictive}. Spotting deviations from the usual behavior is essential in ensuring safety, comfort, and longevity. Thus, the primary task emerges as detecting anomalies relying on collected data by making use of methods that learn from observations, namely the machine learning models \cite{lei2020applications}.  

Anomaly detection is a topic that has garnered significant attention across various communication networks, with IoT networks standing out as one of the most extensively studied domains. However, the distinctive features of the CAN communication protocol set anomaly detection in CAN networks apart from that in IoT networks. A modern vehicle consists of many subsystems containing interconnections between ECUs, sensors, and actuators. The increasing complexity and number of subsystems amplify the difficulty of detecting anomalies and faults within CAN networks. The broadcast type of transmission on the CAN bus and the limited length of the data field in each CAN message make the encryption and authentication of transmitted messages difficult, which makes the CAN network more vulnerable to malicious attacks or intrusions to disrupt message timing and content, which cause abnormal behaviors in different functions of the vehicle. 

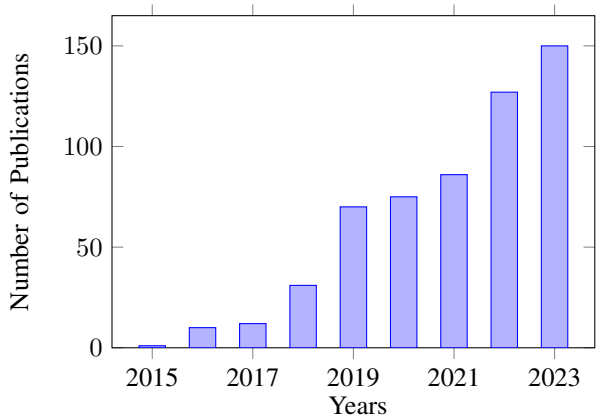
\begin{figure}[!htb]
    \centering
\begin{tikzpicture} 
    \begin{axis} [ybar, 
    ylabel={Number of Publications},
    xlabel={\ Years}, legend style={at={(0.5,0.98)},anchor=north},
    height=6cm, ymin=0, width=8cm, symbolic x coords={2015, 2016, 2017, 2018, 2019, 2020, 2021, 2022, 2023}]
    \addplot coordinates {
        (2015,1) 
        (2016,10) 
        (2017,12) 
        (2018,31) 
        (2019,70) 
        (2020,75) 
        (2021,86) 
        (2022,127)
        (2023,150) 
    }; 

    \end{axis}  
\end{tikzpicture} 
\label{fig:numpub}
\caption{Number of publications on in-vehicle anomaly detection by year in Scopus database}
\end{figure}

Research interest in anomaly detection in in-vehicle networks has increased in recent years. Searching in the abstracts, titles, and keywords in the Scopus database with the query of (("Controller Area Network" OR "CAN Bus" OR "in-vehicle networks") AND ("anomaly detection" OR "intrusion detection")), one can see the increase in the number of conference papers, journal papers and reviews published from 2015 to 2023 in Figure \ref{fig:numpub}.

This study aims to review the techniques and datasets used for anomaly detection in in-vehicle networks. For this purpose, first, we give a general summary of time series anomaly detection and the studies on this subject for the reader to understand the similarities and differences between the approaches applied for time series anomaly detection and CAN bus anomaly detection. The main focus of this survey is the studies for in-vehicle anomaly detection, more specifically CAN bus anomaly detection. To this aim, we selected 85 conference papers and journal articles from the Scopus database, that propose at least one in-vehicle anomaly detection or intrusion detection method and are cited at least once. The survey focuses on specifically CAN bus anomaly detection, other communication environments such as Ethernet are out of the scope. We reviewed and included articles published between 2015 and 2023.

There exist related surveys reviewing in-vehicle anomaly detection. The study in \cite{rajbahadur2018survey} presents a taxonomy of possible cyber-security threads and detection methods for connected vehicles. While its primary objective is to provide a comprehensive framework for understanding the subject matter, it does not specifically delve into anomaly detection techniques.
The survey study in \cite{hu2018review} focuses on security approaches for in-vehicle communication and also gives a brief overview of intrusion detection methods. Similarly, \cite{young2019survey} presents a comprehensive review of intrusion detection systems specifically tailored for CAN bus networks. However, it predominantly addresses vulnerabilities and threats about in-vehicle systems, and the breadth of intrusion detection methods covered in their survey remains somewhat constrained. The studies \cite{lokman2019intrusion, wu2019survey, al2019intrusion, aliwa2021cyberattacks} present a comprehensive examination of intrusion detection methods, with a particular emphasis on attacking strategies and detection algorithms. Notably, these works lack an extensive review of datasets and emerging deep learning techniques, areas that our survey study comprehensively covers. A recent survey \cite{elkhail2021vehicle} gives a review of in-vehicle security issues, vulnerabilities, and malware/intrusion detection methods. The authors provide a comprehensive analysis of malware detection methods. Nevertheless, there is limited information on intrusion detection in this study. Another survey \cite{jo2021survey} on the domain presents CAN bus attacks and countermeasures. It examines the related intrusion detection methods by dividing them into ECU hardware characteristic-based and CAN packet-based methods. Among the CAN packet-based methods, machine learning and deep learning methods are only briefly reviewed. The literature review conducted within the scope of this survey study revealed that a comprehensive survey on anomaly detection for in-vehicle networks is lacking, including a detailed review of datasets and techniques that include both traditional and novel algorithms applied to this problem. We think that this survey will be a useful guide for researchers in industry and academia working on anomaly detection in in-vehicle networks.

The main contributions of this survey can be summarized as follows:
\begin{itemize}
\item A broader review of methods and datasets used for CAN bus anomaly detection is presented. 
\item Both statistical and rule-based methods and novel deep learning techniques are included in the review.
\item Additionally, a summary of time series anomaly detection approaches and datasets is presented to give a broad understanding of the subject.
\end{itemize}

The rest of the paper is organized as follows. Section \ref{section:time-series} describes the overview of anomaly detection on time series, including anomaly detection algorithms and evaluation metrics. A summary of data preparation and machine learning techniques for anomaly detection on the CAN bus is presented in Section \ref{section:anomaly-det-can}. Section \ref{section:public-datasets} gives a summary of public datasets used in anomaly detection research, including both time series and CAN bus datasets. Section \ref{section:overview} discusses the advantages and disadvantages of the methods featured in the previous section. Finally, Section \ref{section:discussion} concludes the paper and mentions future directions in the domain.

\section{Overview on Time Series Anomaly Detection}
\label{section:time-series}

Observations exhibiting an order and correlation in the temporal dimension are referred to as \emph {time series} \cite{esling2012time}. Depending on the sampling, change of temporal values in time series can be even or irregular \cite{li2020learning}. Time series with one variable are called univariate, while the ones across multiple dimensions are termed multivariate, and time series data, as a collection of time series, is usually large and continuously changing across various dimensions \cite{FU2011164}.

The terms anomaly, outlier, and novelty refer to data points that deviate significantly from the rest of the observations and tend to appear interchangeably in the literature \cite{chandola2009anomaly, hawkins1980identification, markou2003novelty}. The phrase \emph{anomaly} will be used throughout the rest of this paper.

Anomalies can be classified as \emph{point anomalies} and \emph{collective anomalies} \cite{al2021review}. Point anomalies are instances where an anomaly can be identified by examining a single data sample in a dataset without contextualizing it with the surrounding points. In time series analysis, point anomalies occur when specific data points exhibit abnormal behavior compared to either the broader series or their immediate neighboring data points. Collective anomalies represent outliers within correlated data samples that are assessed collectively. In the context of time series analysis, collective or subsequence anomalies refer to consecutive data points whose temporal behavior is deemed unusual. For instance, if two successive signal values indicating a vehicle's speed surpass the vehicle's acceleration limit, this would constitute a collective anomaly. When evaluating these signal values individually, they may fall within the vehicle's speed limits, thereby not qualifying as point anomalies.

\subsection{Summary of Techniques Used for Time Series Anomaly Detection}
\label{subsection:techniques-for-ts}

The processing of both univariate and multivariate time series necessitates tailored data preparation steps, which can vary significantly based on the chosen processing approach. For streaming multivariate time series, a common practice involves the application of sliding windows, often with the possibility of overlap, each of fixed length, across the data channels \cite{wolf2018pre}. Alternatively, the fading window strategy that assigns higher weights to more recent observations while diminishing the effects of aging ones, and the landmark window model as an ever-expanding window are employed in the literature \cite{al2021review}.

Concerning univariate time series, point outliers can be detected by parametric statistical thresholding for observed and expected values \cite{blazquez2021review}. Prediction-based methods that use the past data to predict subsequent values such as autoregressive models \cite{williams2003modeling} and deep neural networks \cite{lu2023anomaly} are alternative approaches. These methods calculate the error between predicted and observed values and identify the anomalies based on the error score. Alternatively, distance-based methods \cite{ishimtsev2017conformal} focus on a point's local neighbors, and identify anomalies in case of insufficient neighbors. 

In detecting point anomalies within multivariate time series, one approach involves employing univariate anomaly detection algorithms individually on each time-dependent channel within the multivariate dataset. Alternatively, prediction-based techniques like vector autoregressive models can be extended to a multivariate context \cite{melnyk2016vector}. Another strategy entails utilizing dimensionality reduction methods such as Principal Component Analysis (PCA) to derive a low-dimensional representation of the data. Subsequently, deviations are identified by comparing the reconstructed low-dimensional projection with the original observed time series. An alternative reconstruction-based method involves employing autoencoders \cite{7727309} to learn representations within the multivariate data. Similar to univariate techniques, multi-dimensional distance-based approaches \cite{angiulli2002fast} are also suggested for detecting anomalies in multivariate time series data.
 
Discovering collective anomalies introduces a more challenging problem due to the variable length and periodicity of the sequences \cite{blazquez2021review}. Hence, complex feature extraction techniques are required to end up with powerful representations of the sequences. For univariate series, the simplest approach is to extract subsequences from a series to identify the sequences that are dissimilar to the majority. Using data reflecting normal behavior, anomalies can be detected with defining thresholds. Alternatively, series can be compared to external exemplar time series \cite{jones2016exemplar}. Using past subsequences to predict the future values within a particular horizon to gather prediction errors as anomaly scores is yet another approach \cite{munir2018deepant}. When it comes to detecting collective anomalies in multivariate time series, deep learning algorithms take the lead. 

Anomaly detection methods can be categorized into three learning paradigms: supervised, semi-supervised, and unsupervised approaches. The supervised approach involves mapping inputs to outputs using labeled data, requiring a dataset with annotated input-output pairs. Various classification algorithms are utilized for anomaly detection in time series across different domains. For instance, in the realm of real-time hub motor anomaly detection, multi-layer perceptrons (MLP) have been employed \cite{csimcsir2016real}. Additionally, Convolutional Neural Networks (CNN) have been applied in studies such as \cite{lai2019industrial} and \cite{ullah2021design} for classifying cybersecurity attacks in network traffic data.

In semi-supervised learning, machine learning models are provided with a combination of large amounts of unlabeled data and a small quantity of labeled data, constituting weak supervision. Semi-supervised approaches for time series anomaly detection typically involve training a model exclusively on normal data, with anomalies discerned as patterns diverging from this norm.
For instance, \cite{li2019mad} adopts Generative Adversarial Networks (GANs), where the generator learns the distribution of a multivariate time series solely from normal data, generating reconstructed samples. Simultaneously, the discriminator is trained to differentiate between normal and anomalous data.
Similarly, autoencoders have been utilized in semi-supervised anomaly detection strategies. \cite{nicolau2018learning} and \cite{tagawa2015structured} employ autoencoders on normal data to construct a lower-dimensional feature representation known as the latent space, which is anticipated to exhibit pronounced variations in anomalous instances.

Unsupervised techniques, such as Fuzzy c-Means (FCM), are employed in fault diagnosis for automobile suspension systems \cite{wang2014data}. These methods initially assume the presence of only a cluster representing normal instances. As new data points are introduced, this cluster expands or splits based on a predefined threshold of distance to fault lines connecting existing cluster centers. This approach allows for the detection of anomalies without prior knowledge of fault instances.

Moreover, recent advancements in federated learning \cite{rey2022federated, nguyen2019diot} and homomorphic encryption \cite{alabdulatif2017privacy} for anomaly detection on IoT data highlight the paradigm of processing data in a distributed manner without knowledge of its content.

\subsection{Evaluation Metrics}
\label{subsection:eval}

Anomaly detection is often framed as a binary classification problem, where the negative class encompasses normal cases and the positive class encompasses anomalous cases. To evaluate the performance of an anomaly detector, common metrics used in classifier evaluation can be employed. However, since anomaly detection datasets typically consist of unbalanced data, with normal samples far outnumbering anomalous ones, relying solely on a single metric like accuracy can be misleading. Therefore, in anomaly detection studies, multiple metrics are usually assessed rather than just one, to provide a more comprehensive evaluation of the detector's performance.

Where $TP$, $FP$, $TN$, and $FN$ represent true positive, false positive, true negative, and false negative respectively, definitions of common classification metrics are listed below:

\textbf{Accuracy:} Accuracy is defined as the ratio of the number of correct predictions and the total number of predictions, as given in Equation \ref{eqn:accuracy}.
\begin{equation}
\label{eqn:accuracy}
Accuracy=\frac{T P+T N}{T P+T N+F P+F N}
\end{equation}

\textbf{Precision:} Precision defines how many of the positive predictions are correct, as given in Equation \ref{eqn:precision}.
\begin{equation}
\label{eqn:precision}
Precision=\frac{T P}{T P+F P}
\end{equation}

\textbf{Recall:} Recall, also called sensitivity and True Positive Rate (TPR), defines how many of the positive cases are correctly predicted, as given in Equation \ref{eqn:recall}. 
\begin{equation}
\label{eqn:recall}
Recall=\frac{T P}{\text { Actual Positive }}=\frac{T P}{T P+F N}
\end{equation}

\textbf{F-score:} F-Score is defined as the weighted harmonic mean of precision and recall, where $\beta$ is a real positive factor indicating the weight of recall in the formula, as given in  Equation \ref{eqn:fsc}.
\begin{equation}
\label{eqn:fsc}
F_{\beta}=\left(1+\beta^{2}\right) \times \frac{\text { precision } \times \text { recall }}{\left(\beta^{2} \times \text { precision }\right)+\text { recall }}
\end{equation}

\textbf{False Negative Rate:} $FNR$ defines how many anomalies the detector missed on average, as given in Equation \ref{eqn:fnr}.
\begin{equation}
\label{eqn:fnr}
F N R=\frac{F N}{\text { Actual Positive }}=\frac{F N}{T P+F N}
\end{equation}

\textbf{False Positive Rate:} $FPR$, also called false alarm rate, defines how many positive predictions of the detector were wrong on average, as given in Equation \ref{eqn:fpr}. It is desirable to have a low false alarm rate in anomaly detection.
\begin{equation}
\label{eqn:fpr}
F P R=\frac{F P}{\text { Actual Negative }}=\frac{F P}{T N+F P}
\end{equation}

\textbf{True Negative Rate:} $TNR$, also called specificity, defines how many negative predictions of the detector are correct on average, as given in Equation \ref{eqn:tnr}. 
\begin{equation}
\label{eqn:tnr}
T N R=\frac{T N}{\text { Actual Negative }}=\frac{T N}{T N+F P}
\end{equation}

\textbf{AUC:} AUC, short for Area Under the Receiver Operating Characteristic (ROC) Curve, is a metric used to assess a model's ability to discriminate between classes. It quantifies the overall performance of the classifier by calculating the area under the ROC curve, with values ranging from 0 to 1. A higher AUC value indicates superior classifier performance.

\section{Anomaly Detection on CAN Bus}
\label{section:anomaly-det-can}

Anomaly detection within in-vehicle networks holds significant relevance in the context of condition-based predictive maintenance facilitated by machine learning within the automotive industry. Predictive maintenance, a field extensively explored in the literature, revolves around real-time monitoring of vehicle equipment status \cite{theissler2021predictive}.

Intrusion detection emerges as a critical application of anomaly detection within in-vehicle networks, especially considering the vulnerabilities inherent in the CAN bus. Various forms of intrusions, such as unexpected packet insertion, ECU removal, or corruption, pose significant risks, potentially resulting in critical errors and loss of vehicle control. Consequently, there is a pressing need for systematic research efforts to address these concerns.

The existing literature extensively covers various approaches to anomaly detection within in-vehicle networks, categorizing them based on data preparation procedures and employed methodologies, as summarized in Table \ref{tabone}.  Subsequently, we will delve into a detailed examination of data preparation techniques and anomaly detection approaches in the following sections.

\subsection{CAN Bus Data Preparation for Anomaly Detection}
\label{subsection:data-prep}

\begin{figure}[!htb]
    \center
    \includegraphics[width=90mm]{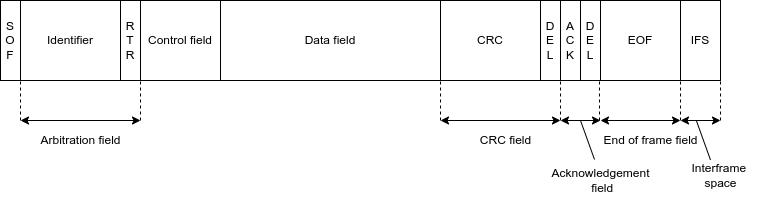}
    \caption{\label{fig:can} CAN Frame Structure}       
\end{figure}

Traffic logs of CAN messages consist of timestamps and CAN packet contents.  Over the CAN bus, four types of frames are exchanged: Data Frames to exchange data, Remote Frames to request transmissions of particular IDs, Error Frames to indicate bus errors, and Overload Frames to inject delays between Data or Remote Frames \cite{d2020cluster}. Concerning anomaly detection, only successfully transmitted CAN data frames are of interest. They comprise several fields \cite{song2020vehicle}, of which the most relevant ones are the arbitration field and the data field. The arbitration field holds the data for identification (ID) and the data field contains the actual vehicle data, which can be of length up to 8 bytes as specified by the Data Length Code (DLC), which is part of the control block. The lower the CAN ID, the more prioritized it is, and its adjacent field, the remote transmission request (RTR) bit is dominant only when information from another party is required. Transmission errors are indicated in the cyclic redundancy check (CRC) and acknowledgment (ACK) fields. A CAN frame structure is shown in Figure \ref{fig:can}.

ECUs continuously transmit CAN messages related to their functions. Multiple CAN IDs are assigned to each ECU 
and two distinct ECUs usually do not
send packets with the same CAN ID. The information on the ECU origins of messages is not public \cite{choi2018voltageids}. CAN specification of messages differ greatly among the manufacturers, and even the vehicle models, while the CAN data frame structure is standardized \cite{kang2016novel}. Consequently, directly extracting actual readings by inspecting raw log data is not always possible unless manufacturers disclose the necessary proprietary information. However, once the distribution of messages is analyzed, CAN messages tend to contain data associated with constant and multi-value fields, counters, or continuous sensory readings \cite{markovitz2017field}. The underlying semantics can be deciphered to a certain extent via reverse engineering through comparison with On-Board Diagnostics (OBD) responses \cite{song2020discovering}, and LibreCAN is an example software package for the task \cite{pese2019librecan}. Since the direct translation of CAN message contents is not always possible, most methods utilize low-level encoding of messages directly to deduce anomalies.

Using fixed-length overlapping sliding windows, time-independent features are extracted using ID and data payload portions of CAN messages, such as the number of packets and their associated statistical moments, the distance between data contents \cite{taylor2015frequency}, the number of distinct IDs, the sum of DLC fields \cite{kwak2021cosine}, counting vector per ID \cite{kuwahara2018supervised}, and entropy calculation based on the distribution of IDs \cite{tian2017intrusion}.

Some studies tend to focus on ID and data fields of CAN messages separately in data preparation. The content of data fields is often treated as features, extracted from traffic logs at various levels such as bit-level \cite{kang2016novel}, byte-level \cite{nowdehi2019casad, fenzl2020continuous}, or represented as 16-set hexadecimal values \cite{qin2021application} and decimal values \cite{lokman2019deep}. Furthermore, network traffic can be regarded as sequences of IDs represented either as binary or hexadecimal strings \cite{song2021self}. Alternatively, it can be visualized as images constructed from IDs in their raw bit form \cite{song2020vehicle} or one-hot encoded vectors for each hexadecimal digit, stacked on top of each other within predefined windows \cite{seo2018gids}. Traffic data can be filtered by CAN IDs and data payload portions of filtered IDs can be used separately to generate subsequences as input to forecasting problems \cite{taylor2016anomaly} or CAN IDs and data field values can be used as integer-encoded features per record on their own \cite{paul2021artificial}. CAN IDs and respective data fields can also be viewed as one-hot encoded categorical features in conjunction with a time variable that holds the time difference between two successive occurrences per CAN ID \cite{app10155062, berger2018comparative}.

Whenever the meaning of the payload in CAN messages is disclosed to the researcher, particular sensory readings can be grouped to form datasets of multivariate time series of measured variables \cite{li2017poster, theissler2017detecting, guo2019detecting, van2019real}. Moreover, irregularly sampled varying numbers of signal values within the CAN traffic can be directly used within dedicated machine learning components as is \cite{9044377} or as subject to further sampling to transform the signals into regular multivariate time series \cite{novikova2020autoencoder}. Alternatively, signal data is extracted from the physical layer of the CAN bus, such as the voltage differences by using oscilloscopes to be represented as time series. Multiple time-domain and frequency-domain features are extracted on the corresponding signals \cite{choi2018voltageids, liu2021vprofile}. 

\subsection{Anomaly Detection Methods on CAN Bus}
\label{subsection:anomaly-can-techniques}
This section offers a comprehensive overview of the techniques outlined in the taxonomy illustrated in Figure \ref{fig:lit_surv}. Table \ref{tab:methods-summary} provides a summary of high-performing anomaly detection approaches trained and evaluated on public datasets. The table includes the learning methods, datasets used, and evaluation metrics/scores reported in the respective studies. The evaluation scores represent the average performance across all attack types.

\begin{figure*}[!htb]
    \centering
    \scalebox{.75}{
    \tikzset{
        basic/.style  = {draw, text width=3cm, align=center, font=\sffamily, rectangle},
        root/.style   = {basic, rounded corners=2pt, thin, align=center, fill=green!30},
        onode/.style = {basic, thin, rounded corners=2pt, align=center, fill=green!60,text width=2cm,},
        tnode/.style = {basic, thin, align=left, fill=pink!60, text width=10em, align=center},
        xnode/.style = {basic, thin, rounded corners=2pt, align=center, fill=blue!20,text width=3cm,},
        wnode/.style = {basic, thin, align=left, fill=pink!10!blue!80!red!10, text width=6.5em},
        edge from parent/.style={draw=black, edge from parent fork right}
    
    }

    \begin{forest} for tree={
        grow=east,
        growth parent anchor=west,
        parent anchor=east,
        child anchor=west,
        edge path={\noexpand\path[\forestoption{edge},->, >={latex}] 
             (!u.parent anchor) -- +(18pt,0pt) |-  (.child anchor) 
             \forestoption{edge label};}
    }
    [Anomaly Detection Methods, basic,  l sep=18mm,
        [Self-supervised \& Unsupervised Methods, xnode,  l sep=18mm,
            [Clustering Methods, tnode]
            [Sequential Methods, tnode] ]
        [Semi-supervised Methods, xnode,  l sep=18mm,
            [Generative Methods, tnode]
            [Reconstructional Methods, tnode] 
            [One-class Methods, tnode] ]
        [Supervised Methods, xnode,  l sep=18mm,
            [Deep Neural Networks, tnode]
            [Sequential Methods, tnode] 
            [Conventional ML Methods, tnode] ]
        [Statistical Methods, xnode,  l sep=18mm,
            [Time Series Analysis, tnode]
            [Fingerprint Characteristics, tnode] 
            [Frequency Analysis, tnode] 
             ] ]
    \end{forest}}
    \caption{A Taxonomy of Anomaly Detection Techniques}
    \label{fig:lit_surv}
\end{figure*}
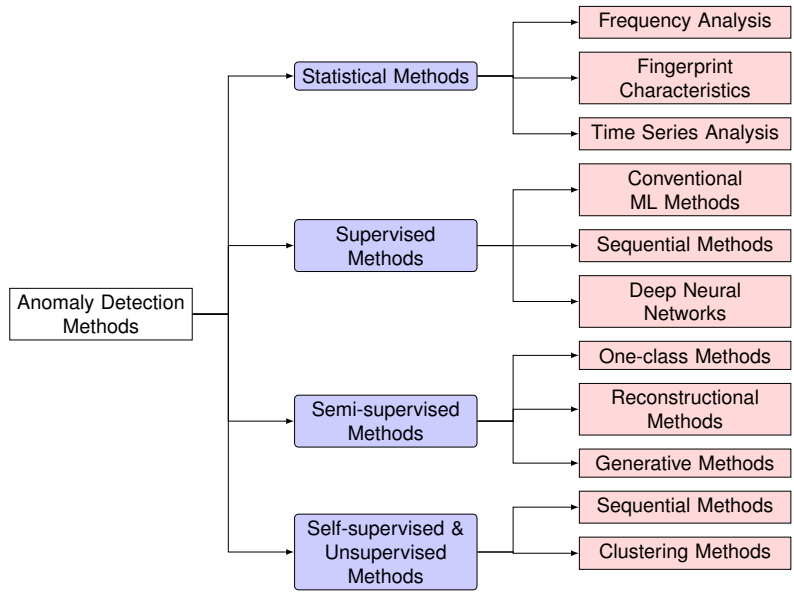

\begin{figure}[!htb]
    \centering
    \scalebox{.65}{
    \begin{tikzpicture}
        \pie{21.68/Unsupervised - Self-supervised,
            22.89/Statistical,
            24.09/Semi-supervised,
            31.32/Supervised}
    \end{tikzpicture}}
    \caption{\label{fig:sup} Distribution of selected publications by learning paradigm}       
\end{figure}
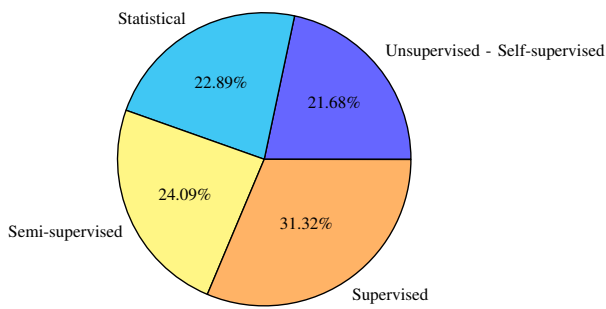

\subsubsection{Statistical Methods}

Frequency-oriented features are extracted over sliding windows that have fixed-length in time, and anomaly detection can be carried out by calculating test statistics to compare window features with historical averages to point out significant differences to be decided as anomalies \cite{taylor2015frequency}. Frequency analysis methods such as wavelet transformation are also utilized for the analysis of behavior profiling of CAN traffic to detect the changes indicating anomalies \cite{bozdal2021winds}. Alternative approaches to detect abnormal patterns in the CAN traffic contain the analysis of time intervals of CAN messages \cite{song2016intrusion, han2021event, yajima2020anomaly}, signal propagation time \cite{murvay2020tidal}, and changes in payload bytes \cite{chevalier2021cyberattack}. 

Several approaches are based on generating fingerprint models using the statistical characteristics of ECUs. One approach proposes calculating clock offsets based on timestamps of CAN messages and predicting the temperature of ECUs at different driving conditions to find the abnormal cases that are different from the fingerprint model of ECUs \cite{tian2020advanced}. Alternative approaches exploit bit time in CAN frames \cite{zhou2019btmonitor}, clock skews \cite{zhou2020clock} and clock drift \cite{ji2018investigating}. 

Statistical thresholding over non-parametric nearest neighbor distances of extracted features \cite{kuwahara2018supervised} is another approach adapted in the literature. Survival analysis is a further statistical method in which CAN messages are divided into chunks and the survival rate of individual CAN IDs in a chunk is measured for thresholding to detect anomalies \cite{HAN201852}. 

Furthermore, time series analysis using statistical modeling algorithms such as Autoregressive Integrated Moving Average (ARIMA) is employed for windowed sequences, whose prediction errors are adapted as anomaly scores \cite{tomlinson2018detection}. Based on the assumption that frequency-related features exhibit certain degrees of similarity in normal traffic streams, the divergence of a window from known patterns can be assessed by cosine similarity to note anomalies \cite{kwak2021cosine}. Extracting finite sequences of IDs that appear on normal traffic data to build a white list of allowed occurrences and comparing them to the patterns encountered in attacks is useful to detect particular anomalies with high precision \cite{markovitz2017field, 9555650}. The proposed methods work well against insertion and deletion attacks that yield the detected anomalies, however, they cannot be generalized to all anomalies. 

When the actual parameters of a vehicle are extracted from CAN messages, domain knowledge and disclosed specifications provided by the manufacturer can be utilized to establish correlated sensory readings. These correlated readings can then be used in a regression setting to predict certain parameters based on the values of others\cite{li2017poster}. The error between observed and estimated values indicates the anomaly score. Whenever the communication matrix is provided, following the signal extraction procedure, static checks based on rules can also be incorporated as an initial processing step to deduce well-known anomalies, which is followed by an application of more sophisticated methods on the extracted features \cite{weber2018hybrid}. An alternative approach proposes checking the abrupt changes in correlations of sensory readings \cite{guo2019detecting}.

\begin{table*}[htb!]
\caption{\label{tab:methods-summary} High-performing models evaluated on public datasets for CAN Bus anomaly detection}
\begin{center}
\scalebox{0.99}{
\begin{tabular}{ |c c c c c c c| } 
 \hline
 \textbf{Method} & \textbf{Learning Approach} & \textbf{Dataset} & \textbf{AUC} & \textbf{Accuracy} & \textbf{F-score} & \textbf{TPR} \\
  \hline

LSTM-ResNet \cite{Song2021ResearchOC} & Supervised & Car Hacking Dataset &  - & 0.91 & 0.91 & 0.91\\
GAN-based GIDS \cite{seo2018gids} & Semi-supervised & Car Hacking Dataset &  0.99 & 0.98 & - & - \\
LSTM-AE \cite{ashraf2020novel} & Semi-supervised & Car Hacking Dataset &  0.97 & 0.99 & 0.99 & 1.00 \\
CLA-DADA \cite{d2020cluster} & Unsupervised & Car Hacking Dataset &  - & 0.99 & - & 0.99 \\
XYF-K \cite{app10155062} &  Unsupervised & Car Hacking Dataset &  - & 0.99 &  0.99 & 0.95 \\

ANN-based AD \cite{paul2021artificial} & Supervised &   OTIDS  & - & 0.99 & 0.99 & 0.99 \\
Entropy-based \cite{wu2018sliding} & Supervised &  OTIDS &  - & 0.96 & - & - \\
SIMATT-SECCU \cite{xiao2019internet} & Semi-supervised &  OTIDS  &  0.94 & 0.99 & - & 1.00\\
MBA-OCSVM\cite{avatefipour2019intelligent} & Semi-supervised & OTIDS & - & - & - & 0.95 \\
CANintelliIDS \cite{javed2021canintelliids} & Supervised &  OTIDS  &  - & - & 0.93 & 0.94 \\

CANet \cite{9044377} & Semi-supervised & SynCAN  &  0.95 & - & - & 0.82 \\
CANShield-Ens \cite{hasan2022canshield} & Semi-supervised & SynCAN &  0.95 & - & - & 0.67 \\
TCN \cite{9555650} & Supervised & SynCAN & - & 0.86 & - & - \\
Ensemble IDS \cite{rajapaksha2023beyond} & Semi-supervised & SynCAN & 0.96 & - & 0.97 & 0.90 \\
INDRA \cite{kukkala2020indra} & Semi-supervised & SynCAN & - & 0.85 & - & - \\
Ensemble IDS \cite{rajapaksha2023beyond} & Semi-supervised & ROAD & 1.00 & - & 1.00 & 0.90 \\
CANShield-Ens \cite{hasan2022canshield} & Semi-supervised & ROAD &  1.00 & - & 1.00 & 1.00 \\
CLAM \cite{sun2021anomaly} & Unsupervised & CAN Signal Extraction \& Translation &  - & - & 0.95 & 0.95 \\
DCNN \cite{song2021self} & Self-supervised & CAN Signal Extraction \& Translation &  - & 0.95 & 0.94 & 0.92 \\

  \hline
\end{tabular}}
\end{center}
\end{table*}

\subsubsection{Supervised Machine Learning Methods}

When multi-variate time series of CAN bus package contents or actual sensory readings are fully labeled for normal and anomalous cases, the problem transforms into a classification setting. To accomplish this classification, model-related heuristics such as anomaly scoring via average path length in an Isolation Forest \cite{hofmockel2018isolation} can be used. Additionally, methods such as entropy analysis with sliding windows \cite{wu2018sliding, yu2020multiple, baldini2020application} and Bloom filters \cite{groza2018efficient} have been explored for anomaly detection in this context. 
As a probabilistic graphical model, Bayesian networks are trained in a supervised manner to classify CAN Bus attacks \cite{pascale2021cybersecurity}. In \cite{tian2017intrusion}, Extreme Gradient Boosting (xgboost) is used as a classifier constructed on decision trees for anomaly detection. Traditional machine learning algorithms such as Random Forests, Support Vector Machines, a mixture of Gaussians and ensemble classifiers of these methods \cite{theissler2017detecting, choi2018voltageids, moulahi2021comparative} are also used in CAN bus anomaly detection.

MLPs are utilized in \cite{kang2016novel} to classify normal and attacked CAN packets. Additional research \cite{fenzl2020continuous, paul2021artificial, kang2016intrusion} also employs MLP-based models for intrusion detection. Furthermore, \cite{park2020hierarchical} introduces a novel multi-labeled hierarchical classification approach to detect anomalies and classify CAN attack types. 
\cite{loukas2017cloud} introduces a supervised model combining Long Short-Term Memory (LSTM) with MLPs. In a similar vein, \cite{zadid9091063} employs LSTMs to develop a binary classifier capable of detecting anomalies within in-vehicle networks by capturing long-term dependencies. Moreover, LSTMs find widespread application in various studies such as \cite{tariq2020can, hossain2020lstm, desta2020long}, focusing on multi-class classification of diverse attack types.

Various studies integrate deep neural networks with other machine learning techniques and data analysis methods to develop robust classifiers.
For instance, \cite{he2021hybrid} introduces a classification model that combines Feed Forward Neural Networks (FFNNs) with cosine-similarity.
Moreover, \cite{wang2020intelligent} introduces an anomaly detection framework comprising two stages: the first stage utilizes statistical characteristics of CAN messages and specific rules, while the second stage incorporates FFNNs.
Additionally, \cite{mehedi2021deep} proposes a LeNet-based model trained with transfer learning for detecting malicious attacks.
In \cite{song2020vehicle}, a CNN-based model utilizing Inception and Residual Networks (ResNet) algorithms is proposed. This model is complemented by a data assembly module that reshapes sequential data into 2D input, enabling direct feeding of CAN bus bit-streams to the network.
RNN and CNN-based modules are often combined for anomaly detection. \cite{javed2021canintelliids} proposes a framework combining GRU and CNN. Similarly,
\cite{Song2021ResearchOC} constructs a classifier using a hybrid architecture of ResNet and LSTM. Here, ResNet is applied to filter time series data and extract distinctive features, while LSTM units are subsequently employed alongside a ResNet module to extract temporal information from the sequences.
Regarding CAN bus intrusion detection, more sophisticated deep learning methods based on global attention on CNNs employing dilated causal convolutions \cite{cheng2022tcan} that operate on CAN message sequences have been proposed.

\subsubsection{Semi-supervised Machine Learning Methods}

To address the challenge of detecting unseen anomalies and effectively handling class imbalance in in-vehicle networks, semi-supervised approaches have been proposed for anomaly detection. 
In these methods, Generative Adversarial Networks (GANs) are repurposed, where the generator is trained to produce synthetic instances resembling normal data. Subsequently, the discriminator utilizes the generated fake data for training instead of specific anomalous data. This approach enables the discriminator to detect unseen anomalies effectively. Several other approaches also explore similar strategies \cite{seo2018gids, abdollah9170806, xie2021threat, yang2021intrusion}.

Another extensively used approach in semi-supervised anomaly detection involves leveraging only normal traffic data within variational autoencoders. In this method, the reconstruction error is adapted as the anomaly score \cite{an2015variational}.
Deep Contractive Autoencoders (DCAE) \cite{lokman2019deep} incorporate an explicit regularizer on the latent representation through the Frobenius norm of the Jacobian matrix of all encoder activation sequences to detect particular attacks on the CAN bus. 
CANet \cite{9044377}, a sophisticated autoencoder architecture, is adapted to CAN Bus traffic data format by feeding induced variable signal values of CAN IDs into dedicated LSTM heads. Next, the size is reduced in the bottleneck layer and all known signals of the traffic are instantaneously reconstructed. By considering multiple CAN messages, the authors have managed to detect even more subtle attacks to a certain extent. Alternatively, by grouping correlated signals within evenly sampled time series, RAEs can be separately trained, whose anomaly scores are combined to yield final predictions \cite{novikova2020autoencoder}. CANnolo \cite{longari2020cannolo}, INDRA \cite{kukkala2020indra}, Ensemble IDS \cite{rajapaksha2023beyond}, and \cite{ashraf2020novel} exploit LSTM autoencoders and Gated Recurrent Unit (GRU) autoencoders.
CANShield \cite{hasan2022canshield} comprises multiple CNN-based autoencoders and employs an ensemble method for the final detection of anomalies on the CAN bus.

Using One-Class Support Vector Machines (OCSVM) in the form of Support Vector Data Description (SVDD) to learn the boundaries of the normal input in a higher dimensional space is another typical semi-supervised method, which categorizes the outsiders as anomalies \cite{theissler2017detecting, tomlinson2021using}. \cite{avatefipour2019intelligent} proposes a model using OCSVM with a modified bat algorithm. Additionally, \cite{kavousi2020effective} introduces a hybrid approach combining OCSVM with wavelet decomposition. Another method proposed for learning irregular patterns in the separating boundaries involves Support Vector Data Description (SVDD) with a Gaussian kernel \cite{9490361}. Moreover, hybrid approaches based on a classification phase by CNNs to identify known faults, followed by Kalman filters with failure detectors that are trained exclusively on normal data are proposed \cite{8684317}.

Transformers, known for their promising performance, especially in Natural Language Processing (NLP) tasks and object detection, have recently been applied in CAN bus anomaly detection as well. In \cite{nam2021intrusion}, a bidirectional Generative Pre-trained Transformer (GPT) is employed to detect anomalies by evaluating the negative likelihood value compared with a predefined threshold.

\subsubsection{Self-Supervised and Unsupervised Machine Learning Methods}

Self-supervised learning has emerged as a powerful paradigm in machine learning, where models are trained to predict certain aspects of the input data without relying on external labels. This approach has gained significant attention in anomaly detection tasks, offering a promising avenue for learning representations directly from data without the need for extensive labeled datasets.

Based on a history of observed values, future values can be forecasted by LSTM in a self-supervised fashion. Predicted sequences are assessed by loss functions that generalize over bit-level binary log loss through applying statistical or logarithmic transformations \cite{taylor2016anomaly, qin2021application}. Associated methods are more flexible in detecting a diverse catalog of known attacks, even though forecasting capabilities are intrinsically limited by the data used in training.

A prominent self-supervised approach involves leveraging normal data exclusively to learn its underlying structures by conditioning an LSTM network to predict the next sequence. Through this process, the trained network generates long CAN ID sequences \cite{song2021self, zhu8673868}. During sequence generation, random noise is introduced via sampling from a Uniform Distribution, resulting in altered sequences known as the noised pseudo-normal instances. After obtaining a sufficient number of instances, a balanced classification is performed using a CNN to detect unseen anomalies.
An alternative approach proposes combining LSTM-RNN with cosine similarity and Pearson correlation \cite{jedh2021detection}. An additional approach, proposed in \cite{sun2021anomaly}, suggests merging one-dimensional convolution with Bidirectional LSTM and an attention mechanism. This fusion aims to predict output signal values and determine anomalous points by evaluating whether the predicted output falls below a certain threshold. Similarly, \cite{kukkala2021latte, xiao2019internet} propose exploiting LSTMs with an attention module. 

To mitigate the challenges posed by the inherent class imbalance problem, researchers have adopted the triplet loss technique, widely used in face recognition within the few-shot learning paradigm, for CAN bus anomaly detection as well. This approach involves mining triplets of anchor, positive, and negative data packets in batches, aiming to minimize the distance between the encodings of positive samples and the anchor while maximizing it between negative samples and the anchor. Consequently, when an anomalous instance is introduced to the network, its learned representation should significantly differ from those of normal instances \cite{app9153174}.

Against more sophisticated attacks that do not alter the prominent characteristics of the in-vehicle network traffic, normal sequences of data payloads are decomposed into a trend, periodic components, and noise via Singular Spectrum Analysis (SSA). The resulting embedding matrix is then further subject to Singular Value Decomposition (SVD) onto which training vectors are projected and clustered to build up a centroid vector that represents normal sequences. Departure from the centroid constitutes the anomaly score for detecting abnormal behavior \cite{DBLP:journals/corr/abs-1909-08407, aoudi2021spectra}.

There exist several unsupervised approaches applied for in-vehicle anomaly detection. Kohonen Self-Organizing Map (SOM) combined with K-means clustering that does not consider class labels to refine the results has been shown to detect more complex attack patterns \cite{app10155062, barletta2020intrusion}. However, the method is constrained by known attack types, training takes a long time, and in case of evolving data retraining from scratch will be necessary. Furthermore, the observed variance in the model performance is quite high, leading to deteriorating performance in particular cases. Another unsupervised algorithm used for anomaly detection is Isolation Forest, which is based on decision trees. It identifies anomalies by recursively dividing the data to isolate outliers from the rest \cite{de2021efficient}. An alternative approach utilizes Hierarchical Temporal Memory (HTM) \cite{wang8274979}, a learning algorithm inspired by the human neocortex. HTM consists of layers arranged hierarchically, enabling the capture of temporal and spatial features from streaming data.

Unsupervised approaches involve Local Outlier Factor \cite{ning2019attacker, tomlinson2021using} and clustering of IDs and related parameters for multiple subsets of normal traffic patterns. These approaches check whether new points fall within the proximity of previously computed legitimate centroids for allowed limits to deduce anomalies at test time \cite{d2020cluster}. Even though this method detects several known attack types with excellence, since it considers the legitimacy of command parameters independently, it might fail to identify the cases in which the attacker changes parameters subtly, for example by copying previous values. Additionally, the performance is heavily affected by hyper-parameters involving cluster sizes and the number of neighbors, which will likely require re-adjustment when changes in normal behavior data are observed in time.  

\subsection{Recent Trends: Data Privacy and Explainability}

Federated learning has gained traction recently, including automotive applications, due to its capacity to train machine learning models across edge devices while preserving data privacy and reducing the need for centralized data storage. Several approaches are proposed to address data privacy using federated learning for anomaly detection on in-vehicle networks. 
\cite{zhang2023federated} utilizes directed attributed graphs from CAN message streams utilizing federated learning.
\cite{hoang2023canperfl} employs personalized federated learning on CAN bus data from various car manufacturers to enhance intrusion detection capabilities by preserving data privacy.
Similarly, \cite{10132145} proposes a privacy-preserving IDS solution for vehicle networks, using federated learning with CNNs to detect intrusions.

With the increasing complexity of AI models, there is a growing demand for transparency and interpretability in AI decision-making processes. Understanding how AI algorithms make decisions in automotive applications is crucial for ensuring trust and accountability. \cite{aziz2022anomaly} applies Explainable Neural Networks (xNN), using K-means clustering to determine feature importance, in intrusion detection. \cite{jeong2023x} improves intrusion detection by translating CAN message payloads into understandable signals, enabling the detection of zero-day attacks without relying on labeled datasets.

\section{Public Datasets for Anomaly Detection Research} 
\label{section:public-datasets}

In this section, we provide a summary of publicly available time series and CAN bus datasets commonly utilized in anomaly detection research. While public CAN bus datasets do exist, it is worth noting that anomaly detection studies on in-vehicle networks often rely on specialized datasets tailored for this purpose. Table \ref{tab:timeseries-data} presents general-purpose time series datasets for anomaly detection, and Table \ref{tab:canbus-data} summarizes CAN Bus datasets for anomaly detection, each briefly described in the following sections.

\subsection{General Time Series Datasets}
\label{subsection:ts-datasets}

In anomaly detection research, various publicly available time series datasets are commonly utilized. One such dataset is the Yahoo Webscope S5 dataset \cite{yahoo}, published by Yahoo Labs. This dataset comprises 367 time series containing anomalies associated with unusual traffic patterns on servers. Numenta Anomaly Benchmark (NAB) \cite{nab} is another multivariate time series dataset having multiple resources containing server metrics, advertisement clicking rates, and traffic data. UNSW-NB15 dataset \cite{unswb} consists of normal and synthesized attack activities in the network traffic. Telemanom \cite{telemanom} is a multivariate time series dataset containing spacecraft anomalies collected by the Soil Moisture Active Passive (SMAP) satellite and the Mars Science Laboratory (MSL). The NASA Turbofan dataset \cite{arias2021aircraft} is constructed by simulating an aircraft engine in different flight conditions and fault modes and reading data from 24 sensors. The dataset comprises 157,523 training and 102,069 testing data points and 10\% of these points represent anomalies or engine failures. WADI \cite{ahmed2017wadi} released by Singapore Public Utility has been constructed by reading 124 sensor measurements during 16 days (14 days under normal conditions, 2 days under attack conditions) from a water distribution system. The dataset contains 104,847 training and 17,270 testing data points. 

\begin{table*}[htb!]
\caption{\label{tab:timeseries-data} Time-series Anomaly Detection Datasets}
\begin{center}
\scalebox{0.99}{
\begin{tabular}{ |c c c c c| } 
 \hline
 \textbf{Dataset} & \textbf{Features} & \textbf{Data Amount} & \textbf{Anomaly Type} & \textbf{Source}  \\
  \hline
UNSW-NB15 \cite{unswb} & \centering{49} & \centering{2,540,038} & Cyber attacks & Network traffic \\
KDD Cup 99 \cite{kddcup} & \centering{41} & \centering{5,209,458} & Cyber attacks & Network traffic \\
NAB \cite{nab} & \centering{58} & \centering{365,551} & Failure & AWS metrics, \newline ad clicks \\
NASA \newline Turbofan \cite{arias2021aircraft} & \centering{24} & \centering{259,592} & Failure & Aircraft engine \\
SMAP \cite{telemanom} & \centering{55} & \centering{429,735} & Unexpected events & Spacecraft engine \\
MSL \cite{telemanom} & \centering{27} & \centering{66,709} & Unexpected events & Spacecraft engine \\
WADI \cite{ahmed2017wadi} & \centering{124} & \centering{122,117} & Cyber attacks & Water distribution \\
Yahoo Webscope \newline S5 \cite{yahoo} & \centering{67} & \centering{572,966 } & Synthetic anomalies & Web traffic \\
\hline
\end{tabular}}
\end{center}
\end{table*}

\subsection{CAN Bus Datasets}
\label{subsection:can-datasets}

While several CAN bus datasets are available for anomaly detection studies, it is important to note that these datasets are primarily designed for vehicle intrusion detection or driver behavior identification. As such, there is a lack of CAN bus data reflecting fault conditions specifically on in-vehicle networks.

The OTIDS dataset \cite{otids} is compiled through the logging of CAN bus message traffic, encompassing a variety of cyber security attacks alongside normal CAN messages. Additionally, a distinct CAN bus traffic dataset has been disseminated by 4TU \cite{4tu}. This dataset comprises CAN log files obtained from two vehicles manufactured by different companies (Opel Astra and Renault Clio) in real-time. It includes instances of malicious messages injected through denial-of-service (DoS) attacks, fuzzing attacks, reconnaissance attacks, as well as normal messages.
Another dataset published by the University of Turku \cite{utu} comprises 180 hours of CAN bus traffic data collected from a Renault Euro VI heavy-duty truck under various traffic conditions, including urban and rural environments. The Car Hacking Dataset \cite{car-hacking} offers approximately 7.5 hours of CAN bus data containing DoS, fuzzy, and spoofing attacks. The Survival Analysis Dataset \cite{HAN201852} includes CAN traffic logs with three types of attacks: flooding, fuzzy, and malfunction. The CAN Signal Extraction and Translation Dataset \cite{song2020discovering} features CAN traffic logs with injected DoS, spoofing, and fuzzy attacks. Additionally, the SynCAN dataset \cite{etas} consists of signals derived from synthetically generated CAN messages, featuring 10 distinct message IDs and a total of 20 signals. Lastly, the ROAD dataset \cite{roaddataset} is a signal-based dataset that includes CAN messages collected over 3.5 hours from a vehicle, encompassing 33 different attacks, including advanced scenarios.

\begin{table*}[htb!]
\caption{\label{tab:canbus-data} CAN Bus Anomaly Detection Datasets}
\begin{center}
\scalebox{0.99}{
\begin{tabular}{ |c c c c c c| } 
 \hline
 \textbf{Dataset} & \textbf{Data Amount} & \textbf{Anomaly Type} & \textbf{Source Type} & \textbf{Real/Synthetic} & \textbf{Source}\\
  \hline

OTIDS \cite{otids} & ~4.6M messages & Vehicle intrusion & Car CAN Bus & Real & 1 vehicle \\

4TU \cite{4tu} & ~3.8M messages & Vehicle intrusion & Car CAN Bus & Real-Synthetic & 2 vehicles \\

UTU \cite{utu} & ~ 530M messages & Vehicle intrusion & Truck CAN Bus  & Real & 1 vehicle \\

Survival Analysis Dataset \cite{HAN201852} & ~1.7M messages & Vehicle intrusion & Car CAN Bus  & Real & 3 vehicles \\

CAN Signal Extraction \& Translation \cite{song2020discovering} & ~5M messages & Vehicle intrusion & Car CAN Bus  & Real & 1 vehicle \\

Car Hacking Dataset \cite{car-hacking} & \centering{~17M messages} & Vehicle intrusion & Car CAN Bus & Real & 1 vehicle \\ 

SynCAN \cite{etas} & \centering{~24h} & Vehicle intrusion & Car CAN Bus & Synthetic &  - \\

ROAD \cite{roaddataset} & \centering{~3.5h} & Vehicle intrusion & Car CAN Bus & Real & 1 vehicle \\
\hline
\end{tabular}}
\end{center}
\end{table*}

\section{Main Findings and Discussion}
\label{section:overview}

Key insights, trends, and conclusions drawn from the reviewed literature on anomaly detection in in-vehicle networks are summarized in the following paragraphs.

\textbf{Training Data.} An essential determinant influencing the selection of the learning paradigm is how the training dataset is curated. The requisite volume of data and labeled samples varies depending on the complexity of the problem at hand. In scenarios where a substantial number of labeled samples are available in the training dataset, supervised learning approaches prove effective in constructing high-fidelity models. However, in cases where the dataset predominantly comprises normal or unlabeled samples, modeling necessitates the utilization of self-supervised, semi-supervised, or unsupervised algorithms. A noteworthy challenge encountered in supervised learning anomaly detection is that the training dataset often covers only certain types of anomalies. Consequently, models trained solely on such data may struggle to detect unseen anomalies effectively.

\textbf{Evaluation Metric.} When designing an anomaly detection solution, the selection of the evaluation metric holds significant importance, as discussed in Section \ref{subsection:eval}. The choice of evaluation metric may vary depending on the application and specific requirements. In scenarios where the cost of overlooking an anomaly outweighs the cost of erroneously classifying a normal instance as an anomaly, prioritizing the rate of correctly classified anomalies can be more critical than minimizing false positives.
If an anomaly detector exhibits a high false positive rate (FPR), it can lead to a substantial number of false alarms, which can be costly and disruptive. Additionally, when there is an imbalance between the number of anomaly and normal samples, assessing the true positive rate (TPR) and true negative rate (TNR) separately, instead of solely relying on accuracy, can provide a more accurate measure of the detector's performance.

\textbf{Task Characteristics.} 
Detecting anomalies within the intricate components of automobiles presents a significant challenge \cite{theissler2017detecting}. Depending on the characteristics of the data and the specific problem at hand, a variety of methods such as statistical approaches, traditional machine learning algorithms, and deep learning techniques may be suitable \cite{GRUNER20201586}. To tackle this diversity of methods effectively, it's crucial to maintain a comprehensive catalog during the model selection process. Furthermore, there is an opportunity to improve detection performance by strategically combining multiple methods, structured according to the hierarchical problem landscape—an area ripe for ongoing research.

\textbf{Risk of Overfitting.} Based on the scores depicted in Table \ref{tab:methods-summary}, it is evident that supervised learning approaches consistently achieve remarkable accuracy and TPR as observed across both the Car Hacking and OTIDS datasets. However, it is important to note that these models are susceptible to overfitting, particularly when trained on small or imbalanced datasets. Overfitting can result in inflated performance metrics on the training set but may lead to poor generalization to unseen data or anomalies that differ from those present in the training set, ultimately resulting in poor real-world performance. More generic representations of normal data can be learned using semi-supervised and self-supervised learning approaches.

\textbf{Resource Limitation.} 
When dealing with limited resources in anomaly detection in in-vehicle networks, various strategies can enhance computational efficiency. These involve choosing lightweight machine learning models, emphasizing feature engineering, ensembling small-size models, applying model compression techniques, and leveraging transfer learning. 

\textbf{Parallelization.}
RNN-based algorithms are structurally suitable for sequence modeling. These models indicate successful performance specifically in collective anomaly detection, as they have the capability of capturing temporal dependencies. However, the disadvantage of these models is that they are unsuitable for parallel processing as they process the data sequentially. Unlike RNN-based models, transformers process sequential data samples all at once rather than sequentially. This offers the advantage of reducing time costs and enabling parallelization. However, transformers face limitations in handling long sequences due to the memory and computational requirements of the self-attention mechanism \cite{lin2022survey}.

\textbf{Data Privacy.} Data privacy is a crucial concern in in-vehicle anomaly detection, as the sensitive nature of vehicular data requires careful handling. Traditional centralized approaches to data processing involve collecting data from multiple sources, which can increase the risk of data breaches. To address these challenges, federated learning has emerged as a promising solution. This approach enables models to be trained across multiple edge devices in a decentralized manner, ensuring that data remains on the device and is not shared with a central server. 

\textbf{Explainability.} Improving transparency in anomaly detection models is essential for fostering trust and understanding in in-vehicle systems. One effective method is to use interpretable machine learning models, such as decision trees, which offer clear insights into how anomalies are detected. Analyzing feature importance further aids in identifying which sensor inputs play a critical role in anomaly detection. Additionally, incorporating attention mechanisms and graph-based architectures can help highlight causal relationships, enhancing the overall understanding of the detection process.

\section{Conclusion}
\label{section:discussion}
This survey undertook a thorough review of both traditional anomaly detection methods and recent innovations in deep learning, offering a comprehensive overview of the current state of the field. This includes an examination of various approaches, ranging from statistical and rule-based techniques to deep neural networks. Among the existing studies, supervised learning stands out as the most frequently used learning paradigm. Additionally, learning methods based solely on normal data are also prevalent in the literature. In this context, RNN and autoencoder-based algorithms are observed to be the most commonly used. However, recently, graph-based algorithms have also gained prominence due to their interpretable nature.

The study also explored the datasets utilized in in-vehicle anomaly detection research.
The reviews conducted throughout the study have highlighted the need for more comprehensive datasets. Most existing public CAN bus datasets primarily include various attacks for intrusion detection. However, there is a significant lack of datasets containing anomalies that are not externally induced but could lead to vehicle malfunctions and faults. 
The insights gained from this review contribute to the broader efforts aimed at improving the performance, reliability, and security of in-vehicle networks. These findings are intended to inform future research directions, guiding the development of more effective and resilient anomaly detection frameworks in automotive systems.

\section*{Acknowledgments}
This study was supported by TÜBİTAK (The Scientific and Technological Research Council Of Türkiye) under the 1501 program with project number 3211202.

{
    \small
    \bibliographystyle{IEEEtran}
    \bibliography{main}
}

\end{document}